\pdfoutput=1

\documentclass[11pt]{article}

\usepackage[]{acl2023}

\usepackage{times}
\usepackage{latexsym}

\usepackage[T1]{fontenc}

\usepackage[utf8]{inputenc}

\usepackage{microtype}

\usepackage{inconsolata}

\usepackage{icomma}
\usepackage{microtype}
\usepackage[ruled, noend]{algorithm2e}
\SetKwInput{KwInput}{Input}                
\SetKwInput{KwOutput}{Output}              

\usepackage[english]{babel}
\usepackage{todonotes}

\newcommand{\Nuno}[1]{}

\usepackage{tcolorbox}

\usepackage{pgfplots}
\usepackage{pgfplotstable}
\pgfplotsset{compat=newest}

\title{Looking for a Needle in a Haystack:\\ A Comprehensive Study of Hallucinations in Neural Machine Translation}

\author{Nuno M. Guerreiro$^{1,2}$ \quad\,\,\, Elena Voita$^{4}$ \quad\,\,\, André F. T. Martins$^{1,2,3}$ \\ \\
$^1$Instituto de Telecomunicações, Lisbon, Portugal \\
$^2$Instituto Superior T\'ecnico \& LUMLIS (Lisbon ELLIS Unit), Lisbon, Portugal\\
$^3$Unbabel, Lisbon, Portugal \quad \quad
$^4$University of Edinburgh, Scotland\\
\footnotesize{\texttt{\{nuno.s.guerreiro, andre.t.martins\}@tecnico.ulisboa.pt}} \quad \texttt{lena-voita@hotmail.com}}

\usepackage{amsmath}
\usepackage{amssymb}
\usepackage{multirow}
\usepackage{tabularx}
\usepackage{booktabs}
\usepackage{colortbl}
\usepackage{xcolor, soul}
\definecolor{HallRedShade}{HTML}{fdf1ec}
\definecolor{HallRedText}{HTML}{b76039}
\sethlcolor{HallRedShade}

\pgfplotsset{
  base/.style={
    ymin=0,
    xtick=data,
    nodes near coords,
  }
}

\pgfplotsset{
  random/.style={
    ybar=3pt,
    bar width=10pt,
    enlarge x limits=0.08,
    tick label style={font=\small},
    every axis y label/.style={at={(-0.06,0.5)},rotate=90},
    every node near coord/.append style={font=\scriptsize},
    every axis legend/.append style={font=\small},
    legend style={at={(0.5,-0.2)},
      anchor=north,legend columns=-1, style={font=\small}},
    legend style={/tikz/every even column/.append style={column sep=0.4cm}}
  }
}

\usepackage{graphicx,subcaption}
\definecolor{new_green}{rgb}{0.0, 0.5, 0.0}

\begin{document}
\maketitle
\begin{abstract}

Although the problem of hallucinations in neural machine translation (NMT) has received some attention, research on this highly pathological phenomenon lacks solid ground. 
Previous work has been limited in several ways: it often resorts to artificial settings where the problem is amplified, it disregards some (common) types of hallucinations, and it does not validate adequacy of detection heuristics. 
In this paper, we set foundations for the study of NMT hallucinations. First, we work in a \textit{natural} setting, i.e., in-domain data without artificial noise neither in training nor in inference. Next, we annotate a dataset of over 3{.}4k~sentences indicating different kinds of critical errors and hallucinations. Then, we turn to detection methods and both revisit methods used previously and propose using glass-box uncertainty-based detectors. Overall, we show that for preventive settings, (i)~previously used methods are largely inadequate, (ii)~sequence log-probability works best and performs on par with reference-based methods. Finally, we propose~\textsc{\mbox{DeHallucinator}}, a simple method for alleviating hallucinations at test time which significantly reduces the hallucinatory rate. 



\end{abstract}


\section{Introduction}

Neural machine translation (NMT) is becoming increasingly  accurate \citep{transformer_vaswani,akhbardeh-EtAl:2021:WMT}, particularly in high resource language pairs where parallel data is abundant. However, even the best systems available today may generate \textit{hallucinations}. These are extremely pathological translations that contain content that is unfaithful to the source sequence. Critically, a tiny fraction of these mistakes is all it takes to compromise user trust or safe deployment of NMT models in production.

Unfortunately, although the problem of hallucinations received some attention, research on this highly pathological phenomenon lacks solid ground. First, previous work used multiple and often overlapping definitions and categories of hallucinations which makes it hard to draw connections between observations made in different works~\citep{lee2018hallucinations,raunak-etal-2021-curious, zhou-etal-2021-detecting}. Next, since hallucinations are extremely rare, previous work focused on settings in which the phenomenon is amplified, e.g. perturbing data either in training or at inference, or evaluating under domain shift \citep{lee2018hallucinations,raunak-etal-2021-curious,muller-etal-2020-domain, wang-sennrich-2020-exposure,voita-etal-2021-analyzing,muller-sennrich-2021-understanding, zhou-etal-2021-detecting}. Critically, the analysis on these works mostly relied on the adequacy of the automatic hallucination detection methods they proposed. However, it is not immediate whether these methods translate well to unperturbed settings.

\begin{table*}[t]
\centering
\scriptsize
\renewcommand\arraystretch{0.85}
\begin{tabular}{
>{\arraybackslash}m{1.0cm} >{\arraybackslash}m{4.8cm}
>{\arraybackslash}m{4.5cm}
>{\arraybackslash}m{4cm}}
\toprule
\textbf{Category} &  \textbf{Source Sentence} & \textbf{Reference Translation} & \textbf{Hallucination}\\ \midrule
Oscillatory & Ist ein Kompromiss aufgrund des zugrundeliegenden Regelsystems unmöglich, so spricht man von Aporie. & The case where, based on the pertinent system of regulations a compromise is not possible, is referred to as Aporia. & Aporia \textcolor{HallRedText}{\hl{is the name of aporia}} , which \textcolor{HallRedText}{\hl{is the name of aporia}}. \\[1.25ex] \midrule
Strongly Detached & Tickets für Busse und die U-Bahn ist zu teuer, vor allem in Stockholm. & Tickets for buses and the subway is too expensive, especially in Stockholm. & \textcolor{HallRedText}{\hl{The hotel is located in the centre of}} Stockholm, \textcolor{HallRedText}{\hl{close to the}} train \textcolor{HallRedText}{\hl{station}}. \\[1.25ex] \midrule
Fully \ \ \ \ \ \  Detached & Die Zimmer beziehen, die Fenster mit Aussicht öffnen, tief durchatmen, staunen. & Head up to the rooms, open up the windows and savour the view, breathe deeply, marvel. &  \textcolor{HallRedText}{\hl{The staff were very friendly and helpful.}} \\ 
\bottomrule
\end{tabular}
\caption{Examples of hallucination types. Hallucinated content is shown \textcolor{HallRedText}{\hl{shaded}}.}
\label{Tab:hallucinationexamples}
\end{table*}

In this work, we set foundations for the study of NMT hallucinations. We take a step back from previous work and, instead of considering perturbed settings for which hallucinations are more frequent, we consider a \textit{natural scenario} and face the actual problem of identifying a small fraction of hallucinations (a ``needle'') in a large number of translated sentences (a ``haystack''). Then, we provide a rigorous comparison among hallucination detection methods. Apart from analysing those proposed in previous work (e.g., heuristics based on anomalous encoder-decoder attention), we also propose to use simple model uncertainty measures as detectors. For each of these methods, we select examples marked as hallucinations, put them together, and gather human annotations. As a result, we introduce a corpus of 3415 structured annotations for different NMT pathologies and hallucinations. We use this corpus for analysis and show that, in preventive settings where high recall is desirable, previously proposed methods are mostly inadequate, and filtering according to standard sequence log-probability performs the best. In fact, it performs on par with the state-of-the-art \textsf{COMET} \citep{rei-etal-2020-comet} which uses reference translation and thus cannot be used in most real-world on-the-fly applications. Surprisingly, its reference-free version \textsf{COMET-QE} \citep{rei-etal-2020-unbabels}, which was shown to generally perform on par with \textsf{COMET}~\cite{toshiptom,freitag-etal-2021-results}, substantially fails to penalise the severity of hallucinations. Overall, methods targeting the phenomena are largely unfit, quality estimation systems fail, and sequence log-probability, i.e. a byproduct of generating a translation, turns out to be the best.

Apart from our analysis of detection methods, we propose \textsc{DeHallucinator}, a method for alleviating hallucinations at test time. At a high level, we first apply a lightweight hallucination detector and then, if a translation is flagged, we try to overwrite it with a better version. For this, we generate several MC-dropout hypotheses~\citep{gal-mcdropout}, score them with some measure, and pick the highest-scoring translation as the final candidate. With this approach, the proportion of correct translations among the ones flagged by the detector increases from $33\%$ to $85\%$, and the hallucinatory rate decreases threefold.

Overall, we show that (i) in preventive settings, previously proposed hallucination detectors are mostly inadequate; (ii) quality estimation techniques fail to distinguish hallucinations from less severe errors; (iii) sequence log-probability is the best hallucination detector and performs on par with reference-based \textsf{COMET}; and, (iv) our \textsc{DeHallucinator} significantly alleviates hallucinations at test time.

Additionally, we release our annotated dataset
along with the model, training data, and code.\footnote{All these resources are available at \url{https://github.com/deep-spin/hallucinations-in-nmt}.}

\section{Taxonomy of Translation Pathologies}
\label{sect:taxonomy}
Choosing a good taxonomy is a compromise between simplicity (which minimizes annotation effort) and comprehensiveness. Thus, generic quality assessment taxonomies, such as MQM~\cite{mqm}, might be unfit or too complex when we focus only on critical errors and hallucinations. For hallucinations, in turn, previous work used multiple, often overlapping, definitions~\citep{lee2018hallucinations, raunak-etal-2021-curious, zhou-etal-2021-detecting, salted_raunak2022}. The taxonomy we build here is rather general: it covers categories considered previously~\cite{lee2018hallucinations,raunak-etal-2021-curious} and others not reported before. For a broader discussion on the taxonomy of hallucinations in NMT, and how it differs from other natural language generation tasks, refer to~\citet{https://doi.org/10.48550/arxiv.2202.03629}.

\subsection{Hallucinations}

To distinguish hallucinations from other errors, we rely on the idea of detachment from the source sequence.  From this perspective, other critical errors such as mistranslation of named entities are not considered as hallucinations. In Section~\ref{sec:high_level_dataset} and Appendix~\ref{sec_app:analysinglesssevere}, we show that properties of hallucinations differ a lot from these other errors and thus our taxonomy is very reasonable.

\paragraph{Oscillatory hallucinations.} These are inadequate translations that contain erroneous repetitions of words and phrases.

\paragraph{Largely fluent hallucinations.} These are largely fluent translations that are unrelated to the content of the source sequence. Previous work assumed they always bear \textit{no relation at all} to the source content~\citep{lee2018hallucinations, raunak-etal-2021-curious}. However, we find that a large proportion of fluent hallucinations partially support the source. Therefore, we also consider severity of a hallucination and distinguish translations that are \textit{fully} detached from those that are \textit{strongly} (but not fully) detached. 

Note that oscillatory hallucinations can also be either fully or only partially detached, but since these hallucinations are less frequent, in what follows we do not split them by severity. We show examples of these hallucination types in~Table~\ref{Tab:hallucinationexamples}.

\subsection{Translation errors}

\paragraph{Undergeneration.}  These are incomplete translations that do not cover part of the source content. This problem is often studied in isolation~\cite{koehn-knowles-2017-six,stahlberg-byrne-2019-nmt,calibration-end-dec-nmt}. Undergenerations are sometimes considered as hallucinations~\cite{lee2018hallucinations} but we do not consider them so in our work.

\paragraph{Mistranslation of named entities.} Appropriately translating named entities (e.g. names, dates, etc.) is also a known difficulty of NMT systems~\citep{ugawa-etal-2018-neural, ne_nmt_mistranslation, hu-etal-2022-deep}. Note that for production systems, this error is rather critical. However, we do not consider it as an hallucination as it does not show detachment from the source but rather an incorrect attempt to translate part of its content.

\paragraph{Other errors.} These are other incorrect translations that do not fit the categories above. They may include errors related to part of speech, word order, and others. For an extensive analysis on machine translation errors, refer to~\citet{vilar-etal-2006-error}.

\section{Hallucination Detection Methods}
\label{sec:halls_det}

Approaches to hallucination detection generally aim to find low-quality translations that may also satisfy addi\-ti\-onal constraints. Previous work either relied only on quality filtering~\cite{lee2018hallucinations,raunak-etal-2021-curious,muller-sennrich-2021-understanding}, or only on heuristics~\cite{berard-etal-2019-naver}, or a combination of the two~\cite{raunak-etal-2021-curious}. We stick to this general form and consider different quality filters and heuristics.

\subsection{Quality Filters}

Quality filters come in two forms: reference-free and reference-based filters. The latter rely on reference translations, while the former do not.


\paragraph{Reference-free methods.} We use the state-of-the-art \textsf{COMET-QE}~\citep{rei-etal-2020-unbabels} for its superior performance compared to other metrics~\citep{mathur-etal-2020-results, freitag-etal-2021-results, toshiptom}. 


\paragraph{Reference-based methods.} Previous work used adjusted \textsf{BLEU} or \textsf{CHRF2} scores of less than~$1\%$ as a standalone criteria~\citep{lee2018hallucinations,raunak-etal-2021-curious,muller-sennrich-2021-understanding,probing_halls_2022}.  
In this work, we analyse \textsf{CHRF2} because it is more suitable for sentence-level evaluation. In addition to this lexical metric, we also consider neural \textsf{COMET}~\citep{rei-etal-2020-comet}, a state-of-the-art reference-based metric~\citep{toshiptom}.

Note that in real-world on-the-fly applications, detecting hallucinations is needed when references are not available. Thus, we use reference-based methods to estimate an upper bound for performance of the other methods.

\subsection{Hallucination Detection Heuristics}
\label{subsec:hall_pred_features}

\subsubsection{Previously Used Heuristics}
\label{subsubsec:prev_used_heuristics}

\paragraph{Binary-score Heuristics.} These heuristics were used by \citet{raunak-etal-2021-curious} to detect oscillatory and fully detached hallucinations. Given a corpus of source-translation pairs, a translation is flagged as an hallucination if it is in the set of~$1\%$ lowest-quality translations, and if:

\begin{itemize}
\item \textbf{Top n-gram count (\textsf{TNG})}. The count of the top repeated $n$-gram in the translation is greater than the count of the top repeated source $n$-gram by at least~$t$ (in their work, $n=4$ and $t=2$);
\item \textbf{Repeated targets (\textsf{RT})}. The translation is repeated for multiple unique source sentences.	
\end{itemize}

\paragraph{Anomalous decoder-encoder attention.} 
Attention patterns\footnote{These patterns are respective to the average of the cross-attention heads of the decoder's last layer.} in which most attention mass is concentrated on the source \texttt{EOS} token are often associated with a model ignoring the source and generating a hallucinatory translation~\citep{lee2018hallucinations, berard-etal-2019-naver, raunak-etal-2021-curious}. We consider two different criteria targeted to find this pattern:

\begin{itemize}
    \item \textbf{\textsf{Attn-to-EOS}}: the proportion of attention paid to the \texttt{EOS} source token;
    \item \textbf{\textsf{Attn-ign-SRC}}: the proportion of source words with a total incoming attention mass lower than $0{.}2$. This was used as a data filtering criterion in \citet{berard-etal-2019-naver}.
\end{itemize}


\subsubsection{Uncertainty-Based Heuristics}

Now we describe the uncertainty measures we propose to use as hallucination detectors. Previously, these were used to improve quality assessments~\cite{fomicheva-etal-2020-unsupervised,zerva-etal-2021-ist}.

\paragraph{Sequence log-probability (\textbf{\textsf{Seq-Logprob).}}}
For a trained model $P(y|x, \theta)$ and a generated translation $y$, \textsf{Seq-Logprob} (i.e., model confidence) is the \textit{length-normalised} sequence log-probability:
\begin{equation}
    \frac{1}{L} \sum\limits_{k=1}^{L} \log P(y_{k} \mid y_{<k}, x, \theta).
    \label{eq:seq_logprob}
\end{equation}
We hypothesise that when hallucinating, a model is not confident.

\paragraph{Dissimilarity of MC hypotheses (\textbf{\textsf{MC-DSim}).}}
This method measures how the original hypothesis~$y$ disagrees with hypotheses $\{h_1, \dots, h_{N}\}$ generated in stochastic passes. For the same source sentence, we generate these new hypotheses  using Monte Carlo (MC) Dropout \citep{gal-mcdropout}. Then we evaluate the average similarity:
\begin{equation}
  \frac{1}{N} \sum_{i=1}^{N}\textsc{sim}(h_{i}, y).
    \label{eq:soh}
\end{equation}
Different similarity measures can be used in place of $\textsc{sim}$ (e.g. METEOR~\cite{banerjee-lavie-2005-meteor}, BERTScore~\cite{Zhang2020BERTScore:}, etc.). We follow previous work and use METEOR with $N=10$~\cite{fomicheva-etal-2020-unsupervised, zerva-etal-2021-ist}.

\subsection{Trained Hallucination Detection Model}
\label{subsec:tokhalmodel}
An exception from the general framework of hallucination detection is the work by
\citet{zhou-etal-2021-detecting} who \textit{learn} to detect token-level hallucinations. Specifically, the authors 
create synthetic data where they randomly corrupt some tokens in a translation and reconstruct them with the BART model~\cite{lewis-etal-2020-bart}. Then, the authors fine-tune a pretrained language model to identify the replaced tokens.

\paragraph{\textbf{\textsf{TokHal-Model.}}} We evaluate the proportion of tokens that are predicted to be hallucinated and use this as a detection score.

\subsection{Binary vs Continuous Scores}
\label{subsec:binary_cont_scores}
The methods above fall into two categories:
\textit{binary-score} and \textit{continuous-score} heuristics.
The former (only \textsf{TNG} and \textsf{RT}) output a value in $\{0, 1\}$, whereas the latter output a value in $\mathbb{R}$ and the prediction is made depending on a chosen threshold. 

\section{Experimental Setting}
\label{sec:experimental_setting}

\paragraph{Model.} We use Transformer base~\citep{transformer_vaswani} from \texttt{fairseq}~\cite{ott-etal-2019-fairseq}.


\paragraph{Data.} We use the WMT2018 DE-EN news translation data excluding Paracrawl~\citep{bojar-etal-2018-findings}~-- 5{.}8M sentence pairs. We randomly choose 2/3 of the dataset for training and use the remaining 1/3 as a held-out set for analysis. For validation, we use the \textit{newstest2017} dataset.

\paragraph{Held-out Data Filtering.} We are mainly interested in hallucinations produced for clean data. Since our held-out data comes from the WMT2018 training dataset and thus can be noisy, we filter it using Bicleaner, the filtering tool used in official releases of filtered ParaCrawl data \citep{prompsit:2018:WMT, prompsit:2020:EAMT,10.1162/tacl_a_00447}. Following previous work, we exclude examples with a score below~$0{.}5$ and end up with about 1.3M examples.

All details on preprocessing, hyperparameters and implementation can be found in Appendix~\ref{appendix:experimental_setup}.

\section{Hallucinations Dataset}
\label{sec:halls_dataset}
To analyse the effectiveness of hallucination detection criteria, we pick a subset of examples that are likely to be hallucinations, and obtain fine-grained annotations from professional translators. 


\subsection{Data for Annotation}

Our data selection is motivated by two goals: 
(i)~find as many hallucinatory translations as possible~-- to analyse hallucinations,
(ii)~pick some translations from a long tail of hallucination detection predictions~-- to analyse the behaviour of these detection methods. Thus, we first pick~250 worst-scored samples for each heuristic and quality filter (including binary assignments obtained through \textsf{TNG} and \textsf{RT}). Next, we turn to a broader set of samples and consider translations whose scores fall below a chosen percentile for a given method, i.e. we consider long tails of the scores.\footnote{From now on, we refer to  examples contained in a long tail of a method as ``flagged'' or ``detected'' by this method.} For in-domain settings, previous work reported hallucinatory rates of~$0{.}2-2\%$. However, these rates were either obtained on noisy and/or low-resource data or using weaker models. Therefore, in our cleaner in-domain setting with a stronger model, we expect the hallucination rate to lie in the lower end of the indicated range. Thus, we consider approximately~$0{.}4\%$ of the worst scores (which amounts to 5000 flagged translations for each criteria).\footnote{The threshold for prediction is consistent with this percentile. For practical details, refer to Appendix~\ref{sec:prac_recommendations}.}  From these, we sample 250 examples and add them to the dataset. In total, we end up with 3415 examples for annotation.\footnote{Note that a sample originally obtained from the worst scores or sampled from the long tail of a given method may belong to the long tail of another method.}

\subsection{Guidelines and Annotation}

The annotation guidelines are developed according to the taxonomy defined in Section~\ref{sect:taxonomy}. All details on data collection can be found in Appendix~\ref{appendix:data_collection}.

\section{High-level Overview of the Dataset}
\label{sec:high_level_dataset}
\subsection{General Statistics}
\begin{figure}[t]
\centering
{\includegraphics[width=\columnwidth]{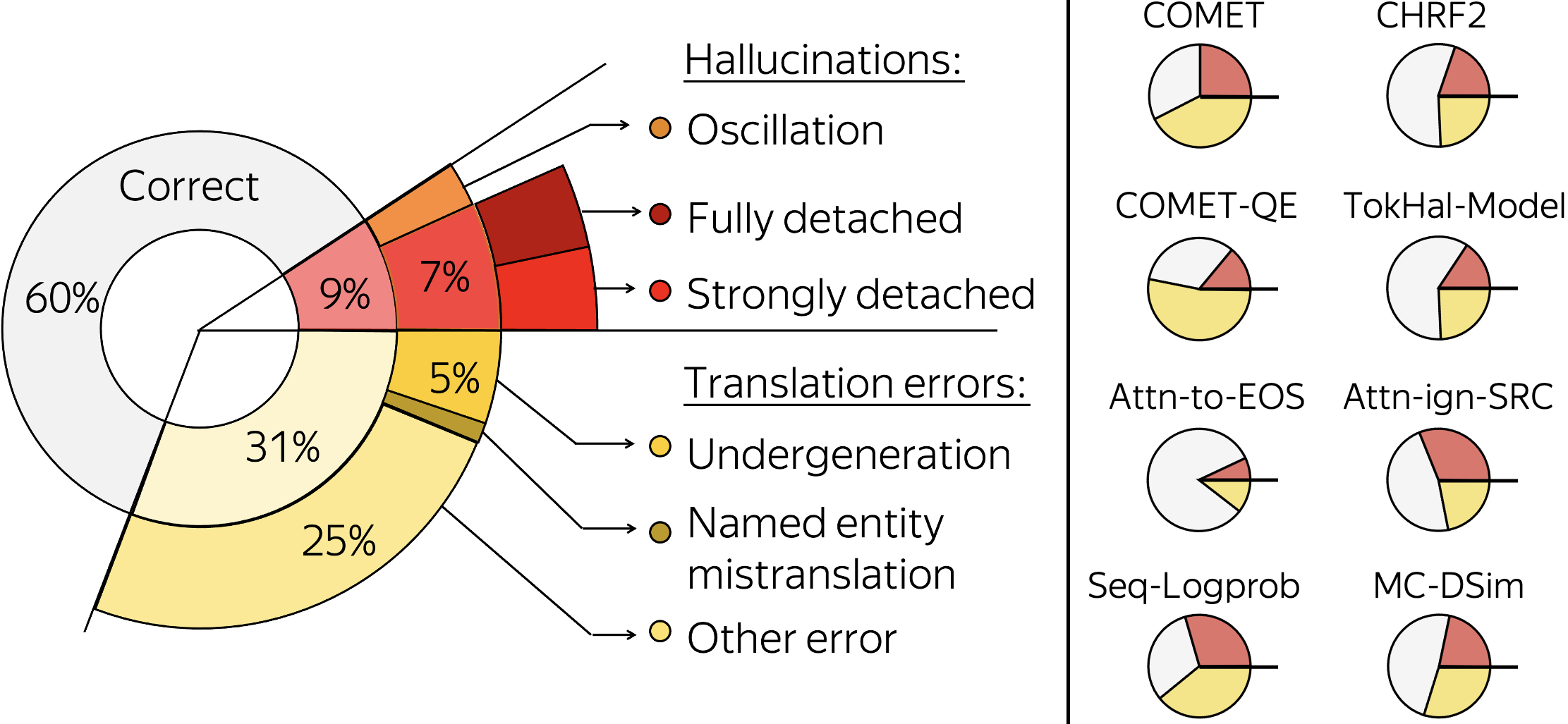}}
\caption{Overall (left) and method-specific (right) statistics of human annotation results. Method-specific statistics show the percentages of correct translations (grey), translation errors (yellow) and hallucinations (red) among the examples flagged by each method.}
\label{fig:dataset}
\end{figure}

Figure~\ref{fig:dataset} gives a structured overview of dataset statistics. First, we see that while we picked translations that are likely to be pathological, $60\%$ of the dataset consists of correct translations. This highlights that with the existing methods,  finding poor translations reliably is still challenging. Next, note that most of the incorrect translations have translation errors that are not severe enough to be deemed hallucinations. This agrees with the view that hallucinations lie at the extreme end of the spectrum of MT pathologies~\cite{raunak-etal-2021-curious}. Finally, the results of the annotation confirm that our data selection is very reasonable. Indeed, while previous work has reported hallucinatory rates of~$0.2-2\%$ in in-domain settings, we see that, for a reasonably numbered collection of examples, our hallucination rate is substantially higher~($9\%$) -- 294 hallucinations among the 3415 translations. In Figure~\ref{fig:dataset}, we also show the method-specific statistics of human annotation results for each heuristic and quality filter. Unsurprisingly, the long tails of each method display different characteristics. For example, almost all translations flagged by \textsf{Attn-to-EOS} are correct, whereas the proportion of good translations flagged by \textsf{COMET-QE} or \textsf{Seq-Logprob} is rather small. We will analyse this further in Section~\ref{sect:analysing_detection_criteria}. 

\begin{figure*}[t]
    \centering
    \begin{subfigure}[b]{0.63\columnwidth}        
        \includegraphics[width=.935\columnwidth]{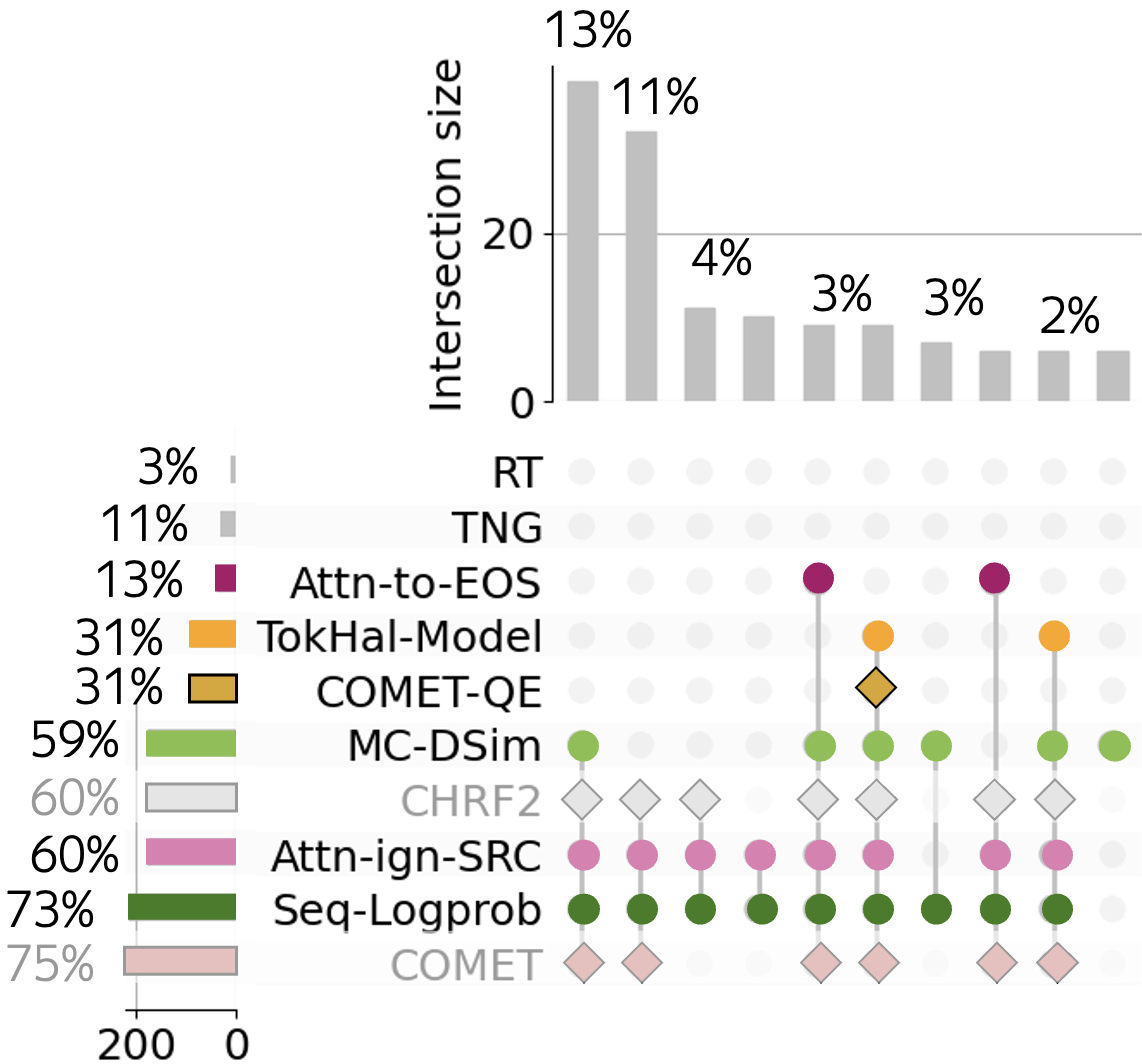}
        \caption{Hallucinations}
        \label{fig:intersections_hal}
    \end{subfigure}
    \ \ \ \ 
    \begin{subfigure}[b]{0.67\columnwidth}        
        \includegraphics[width=.935\columnwidth]{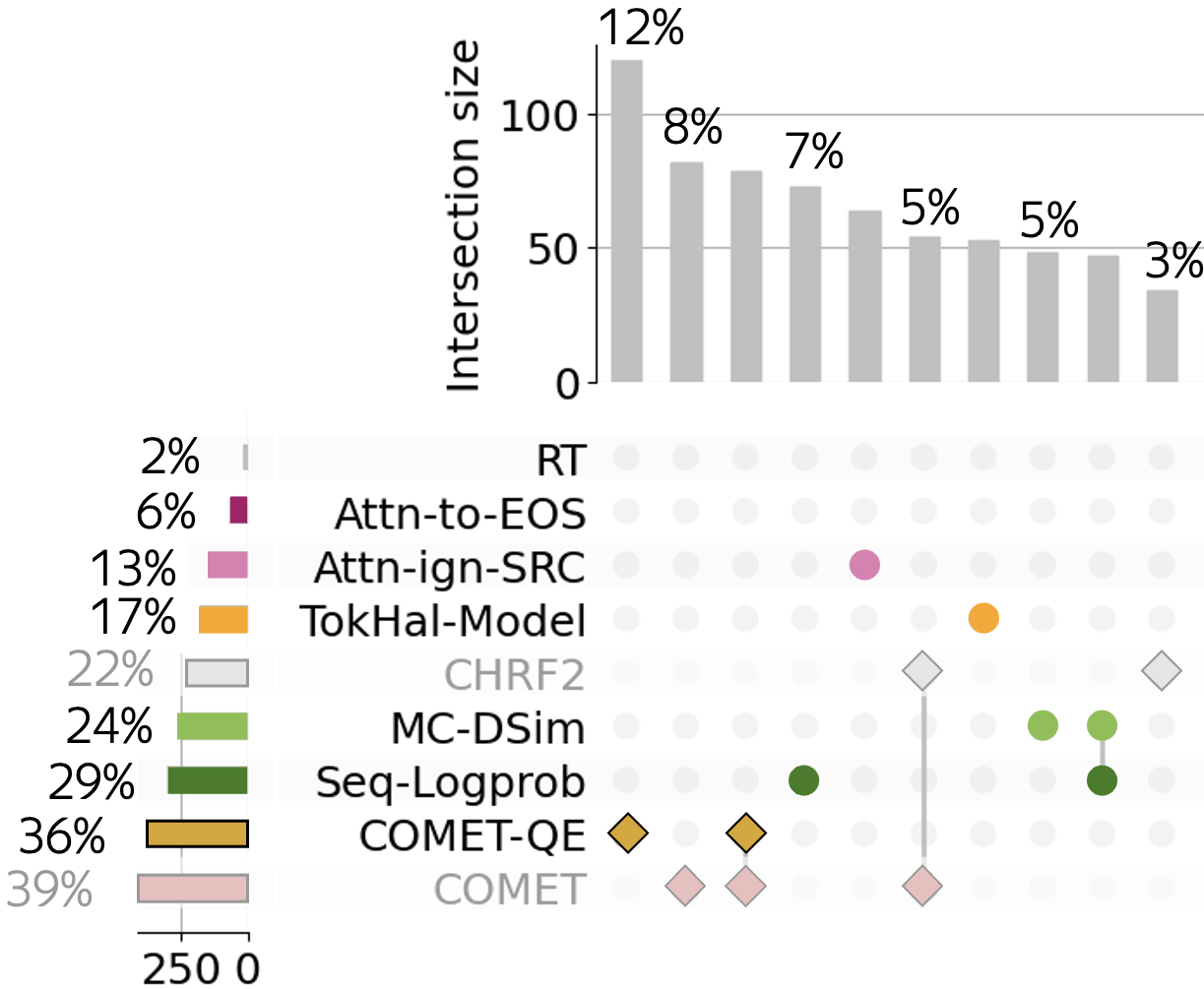}
        \caption{MT errors}
        \label{fig:intersections_errors}
    \end{subfigure}
    \ \ \ \ 
    \begin{subfigure}[b]{0.63\columnwidth}        
        \includegraphics[width=.935\columnwidth]{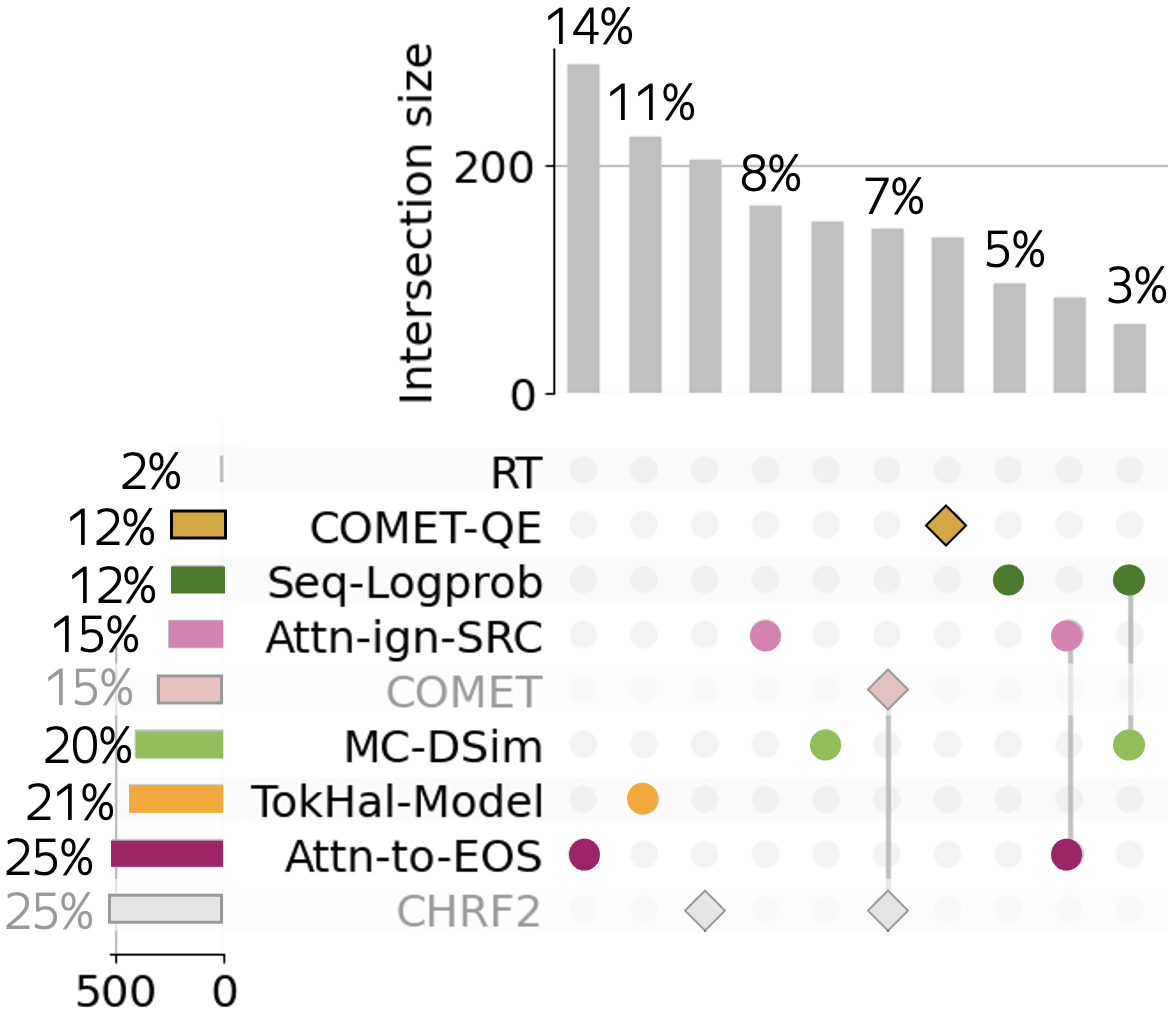}
        \caption{Correct translations}
        \label{fig:intersections_correct}
    \end{subfigure}
    \caption{Structure of the sets of translations flagged by the considered methods. 
    Horizontal bars show the proportion of examples flagged by each method among all translations of the considered category. 
    Each vertical bar shows the size for the set of translations that are (i)~flagged by all the methods marked in the corresponding column and (ii)~not flagged by any of the rest; only the top intersections are shown. Quality filters are shown with diamond marks, and detection heuristics~-- with circles. Methods requiring reference translations are shaded.}
\label{fig:intersections_all}
\end{figure*}

\subsection{MT Errors vs Hallucinations}
Now let us look separately at the sets of examples with hallucinations and other translation errors. Figure~\ref{fig:intersections_all} shows the structures of these sets
with respect to interactions of the different criteria.

We see that it is reasonable to consider hallucinations separately from other errors: patterns in Figures~\ref{fig:intersections_hal} and~\ref{fig:intersections_errors} are substantially different. For example, 
for translation errors, \textsf{COMET-QE} performs well, being on par with reference-aware \textsf{COMET}~(Figure~\ref{fig:intersections_errors}). However, for hallucinations, it identifies less than half the amount of examples identified not only by \textsf{COMET}, but even by simple uncertainty-based heuristics~(Figure~\ref{fig:intersections_hal}). Furthermore, Figure~\ref{fig:intersections_all} reveals a significant difference between interactions of different criteria for hallucinations and for other translations: hallucinations in our data are most often flagged by multiple criteria simultaneously (e.g. \textsf{Seq-Logprob} flags the majority of hallucinations detected with \textsf{Attn-ign-SRC} and \textsf{MC-DSim}), whereas most MT errors and correct translations are flagged by a single method. 
This difference in patterns supports our choice of taxonomy: properties of hallucinations are very different from those of other less severe errors. 

\section{Analysing Detection Criteria}
\label{sect:analysing_detection_criteria}

In this section, we provide a comprehensive analysis of the performance of the heuristics and quality filters introduced in Section~\ref{sec:halls_det}.

\subsection{Quality Filters}
\label{subsec:qualityfilters}
Here we start by analysing reference-based methods, namely \textsf{COMET} and \textsf{CHRF2}, and then turn to the reference-free \textsf{COMET-QE}. Overall, our results show that reference information is helpful, while \textsf{COMET-QE} fails to penalise hallucinations. 


\paragraph{Reference information helps detection.} Figure~\ref{fig:intersections_hal} shows that leveraging reference information helps detecting hallucinations: \textsf{COMET} detects more hallucinations than any of the other methods. As expected, lexical-based \textsf{CHRF2} is significantly worse than neural-based \textsf{COMET}. In fact, it lags behind several heuristics (e.g., \textsf{Seq-Logprob}).

We also explore whether previously proposed methods for detecting hallucinations under domain shift are appropriate in our cleaner in-domain setting. Specifically, we follow~\citet{muller-sennrich-2021-understanding} and consider translations with \textsf{CHRF2} score lower than~$1\%$. Strikingly, for our clean setting, this approach is inadequate: in the entire held-out set of 1.3M examples, it flagged only 2~translations. This suggests that methods suitable for noisy settings (e.g.,  domain shift) might not be applicable in settings where models are less likely to hallucinate. 




\begin{table}[t]
\centering
\footnotesize
\newcolumntype{S}{>{\centering\arraybackslash}m{0.1\textwidth}}
\newcolumntype{N}{>{\centering\arraybackslash}m{0.065\textwidth}}
\newcolumntype{G}{>{\arraybackslash}m{2in}}
\renewcommand\arraystretch{1}
\begin{tabular}{>{\arraybackslash}m{0.08\textwidth} >{\centering\arraybackslash}m{0.065\textwidth} >{\centering\arraybackslash}m{0.075\textwidth}  >{\centering\arraybackslash}m{0.0325\textwidth} >{\centering\arraybackslash}m{0.0325\textwidth} >{\centering\arraybackslash}m{0.0325\textwidth}}
\toprule
\multirow{2}{*}{\!\!\!\textbf{Heuristic}} & \multirow{2}{*}{\textbf{Correct}} & \multirow{2}{0.075\textwidth}{\centering \textbf{MT Errors}} & \multicolumn{3}{c}{\textbf{Hallucinations}} \\
& & & \texttt{OSC} & \texttt{SD} & \texttt{FD}\!\!\!
\\ 
\midrule
\!\!\!\textsf{TNG} & 0 & 0 & 32 & 0 & 0\\
\!\!\!\textsf{RT} & 18 & 19 & 2 & 1 & 7 \\ \midrule
\!\!\!\textsf{All dataset} & 2048 & 1073 & 86 & 90 & 118\!\!\! \\ 

\bottomrule
\end{tabular}
\caption{Translations flagged by \textsf{TNG} and \textsf{RT}.}
\label{tab:binary_heuristics}
\end{table}

\paragraph{COMET-QE fails to penalise hallucinations.} From Figure~\ref{fig:dataset}~(right) we see that, as expected, most of the translations flagged by \textsf{COMET-QE} are incorrect. However, the vast majority of them are not hallucinations. Indeed, Figure~\ref{fig:intersections_hal} shows that 
\textsf{COMET-QE} is one of the worst hallucination detectors, meaning that it fails to rank by the severity of a translation pathology. This supports the hypothesis made in previous work: since quality estimation models are mostly trained on data that lacks negative examples, they may be inadequate for evaluating poor translations~\cite{takahashi-EtAl:2021:WMT, sudoh-etal-2021-translation}.




Overall, among the considered quality filters, only \textsf{COMET} may be used as a hallucination detector. However, it is important to keep in mind that, in on-the-fly applications, detecting hallucinations is needed when references are not available, rendering reference-based methods not applicable.

\subsection{Detection Heuristics}
\label{subsec:detection_heuristics}
The results in Section~\ref{subsec:qualityfilters} leave a relevant gap: where do we turn to when references are not available? Is there any information, besides quality, that may help detecting hallucinations? Our evidence suggests that in preventive settings, previously proposed heuristics are mostly inadequate, and uncertainty may be the answer for the questions we pose.

\paragraph{Binary-score heuristics perform the worst.} Table~\ref{tab:binary_heuristics} shows the number of detected translations for 
Top $n$-gram count~(\textsf{TNG}) and Repeated targets~(\textsf{RT}). These heuristics are targeted to identify oscillatory and fully detached hallucinations, respectively. We see that while \textsf{TNG} obtains perfect precision, it fails to identify more than half of the oscillatory translations. \textsf{RT}, in turn, performs poorly across the board: only a few hallucinations are detected, and a significant proportion of flagged translations turn out to be correct. Moreover, Figure~\ref{fig:intersections_all} shows that even if we join sets of translations detected by \textsf{TNG} and \textsf{RT} (as done in \citet{raunak-etal-2021-curious}), altogether we get \textit{fewer hallucinations than almost any other considered method}. Thus, in preventive settings, these methods are highly inadequate.

\paragraph{Anomalous attention is mostly \textit{not} hallucination.} Figures~\ref{fig:dataset} and~\ref{fig:intersections_all} show that behaviors of \textsf{Attn-to-EOS} and \textsf{Attn-ign-SRC} are significantly different. First, \textsf{Attn-to-EOS} \textit{is not indicative of hallucinations}. Indeed, attention patterns in which most attention mass is concentrated on the \texttt{EOS} token largely correspond to correct translations. On the other hand, \textsf{Attn-ign-SRC} performs well and is second only to uncertainty-based \textsf{Seq-Logprob}. Such a difference in performance is surprising: both methods are motivated by a common belief that if almost all the attention mass is concentrated on the source \texttt{EOS} token, a translation is likely to be a hallucination~\cite{berard-etal-2019-naver,raunak-etal-2021-curious}. In fact, both methods were designed to identify this specific pattern. However, patterns identified with \textsf{Attn-ign-SRC} span from attention mass coming to various uninformative tokens (e.g., punctuation) to examples where attention is
mostly diagonal~(typically, these correspond to undergenerations). We show examples of such attention maps in Appendix~\ref{app:attn_patterns}. Overall, the results highlight a disparity between what is \textit{believed to indicate} hallucinations and what \textit{actually indicates} them. 



Note that while \textsf{Attn-ign-SRC} performs relatively well, it should be used with caution: attention-based heuristics rely on the assumption that attention patterns reflect model reasoning. This assumption is not reliable: although there is evidence that attention can play recognizable roles~\cite{voita-etal-2018-context,voita-etal-2019-analyzing}, a lot of work questions attention explainability~\citep{wiegreffe-pinter-2019-attention, jain-wallace-2019-attention, serrano-smith-2019-attention, bastings-filippova-2020-elephant, pruthi-etal-2020-learning}. Since hallucinations identified by \textsf{Attn-ign-SRC} are overwhelmingly contained among the ones identified by \textsf{Seq-Logprob} (Figure~\ref{fig:intersections_hal}), we recommend using the latter instead.

\paragraph{TokHal-Model is unfit for natural hallucinations.}
Let us recall that the model used for the \textsf{TokHal-Model} scores was trained to identify replaced tokens in corrupted translations that are fluent and do not differ much from the original ones~\cite{zhou-etal-2021-detecting}. This means that during training, the model was unlikely to observe highly pathological translations that reflect the types of hallucinations produced by actual NMT systems. This raises several concerns when using this model in our setting. For example, it might incorrectly flag adequate tokens such as synonyms or paraphrases. What is more, since severely flawed examples are mostly out of distribution for the model, labels predicted for such translations may be unreasonable. 

The results confirm our concerns: Figures~\ref{fig:dataset} and~\ref{fig:intersections_hal} show that (i)~the vast majority of translations flagged by \textsf{TokHal-Model} are correct and (ii)~it is one of the worst hallucination detectors. 

\paragraph{Model confidence may be all you need.} Figure~\ref{fig:intersections_all} shows that \textsf{Seq-Logprob} is \textit{the best heuristic} and performs on par with reference-based \textsf{COMET}.  This means that the less confident the model is, the more likely it is to generate an inadequate translation. This agrees with some observations made in previous work on quality estimation~\cite{fomicheva-etal-2020-unsupervised}. Interestingly, such performance of the method contrasts with its simplicity: \textsf{Seq-Logprob} scores are easily obtained as a by-product of generating a translation. This distinguishes the method from all the rest that require additional computation (e.g., corpus-level search for \textsf{RT} or generating multiple hypotheses for \textsf{MC-DSim}).

On another note, NMT models have been found to be miscalibrated~\citep{calibration-end-dec-nmt}. Thus, investigating the impact of recalibration methods -- aimed at enhancing prediction reliability -- on the detection quality of \textsf{Seq-Logprob} constitutes an interesting avenue for future research.

\paragraph{\textsf{MC-DSim}: hallucinations are mostly unstable.} The intuition behind this heuristic is simple: when faced with a source sentence for which a good translation is not immediate, the set of hypotheses the model ``keeps at hand'' may be very diverse. Indeed, \textsf{MC-DSim} performs relatively well and identifies a good proportion of hallucinations~(Figures~\ref{fig:dataset}, \ref{fig:intersections_all}). Thus, most hallucinations are \textit{unstable}: dissimilarity of MC hypotheses helps to identify them. 

Note that while MC-DSim is not the best choice
for detection, later we will show that the intuition
behind this method is very helpful for alleviating
hallucinations at test time (Section~\ref{sec:reversal}).

\begin{figure*}[t]
    \centering
    \begin{subfigure}[b]{0.3\linewidth}        
        \centering
        \includegraphics[width=0.93\linewidth]{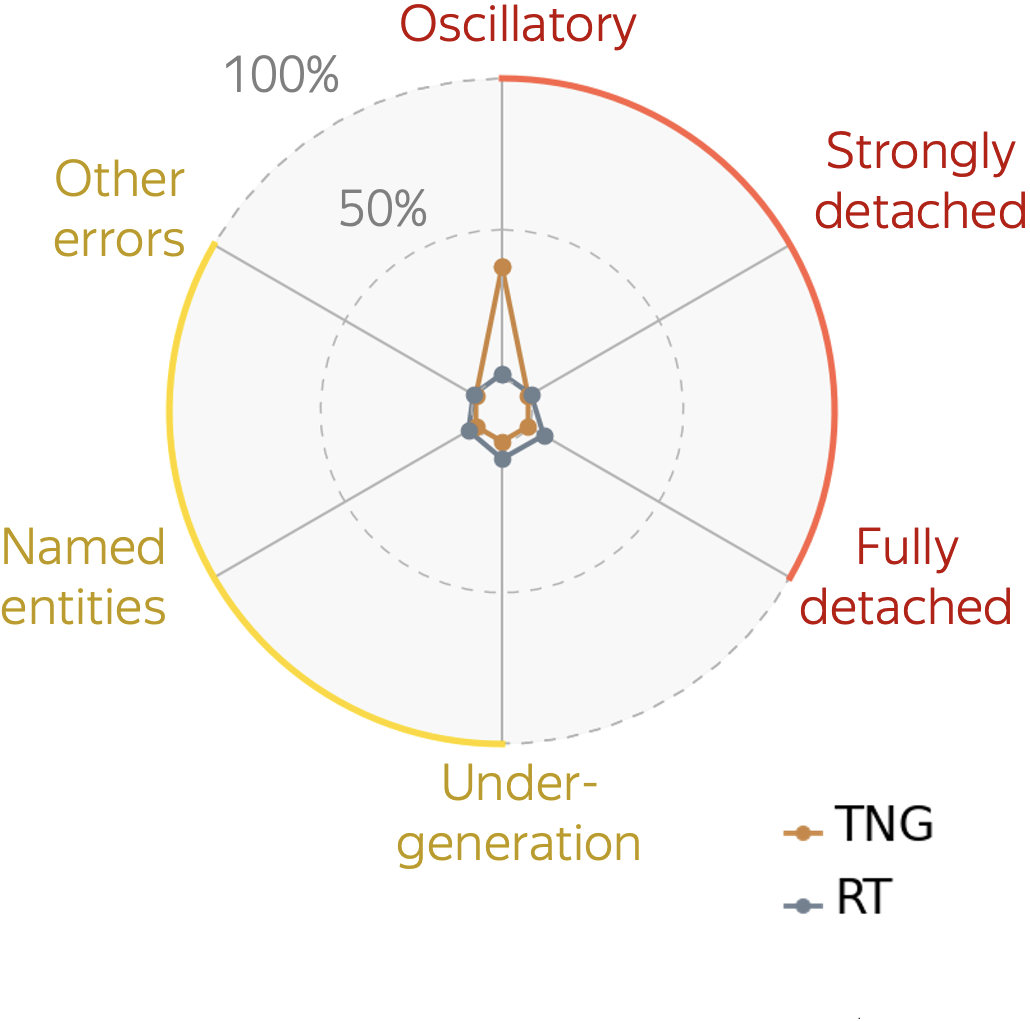}
        \caption{Binary-score heuristics}
        \label{fig:spiderA:heuristics_binary}
    \end{subfigure}
    \ \ \ \
    \begin{subfigure}[b]{0.325\linewidth}        
        \centering
        \includegraphics[width=0.91\linewidth]{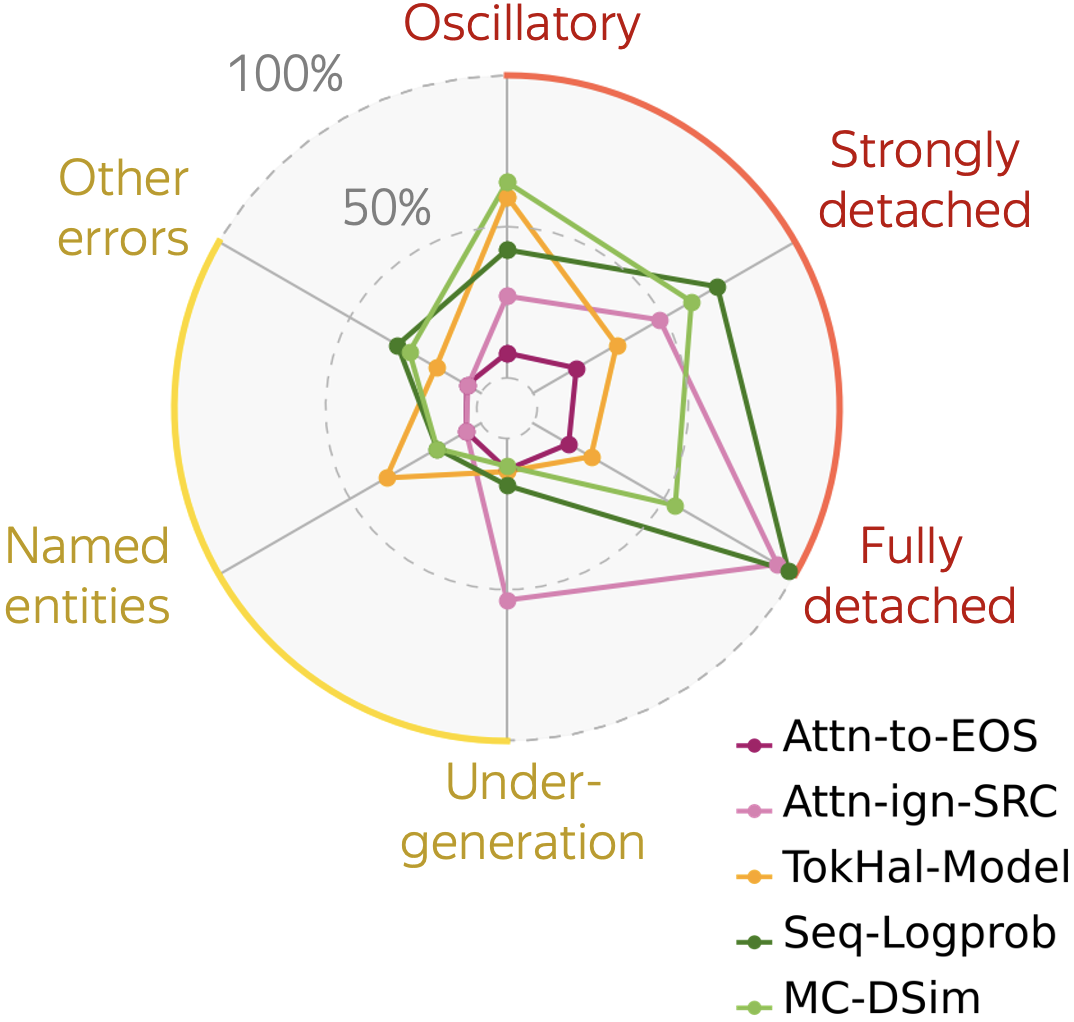}
        \caption{Continuous-score heuristics}
        \label{fig:spiderB:heuristics_continuous}
    \end{subfigure}
    \ \ \ \
    \begin{subfigure}[b]{0.3\linewidth}        
        \centering
        \includegraphics[width=0.95\linewidth]{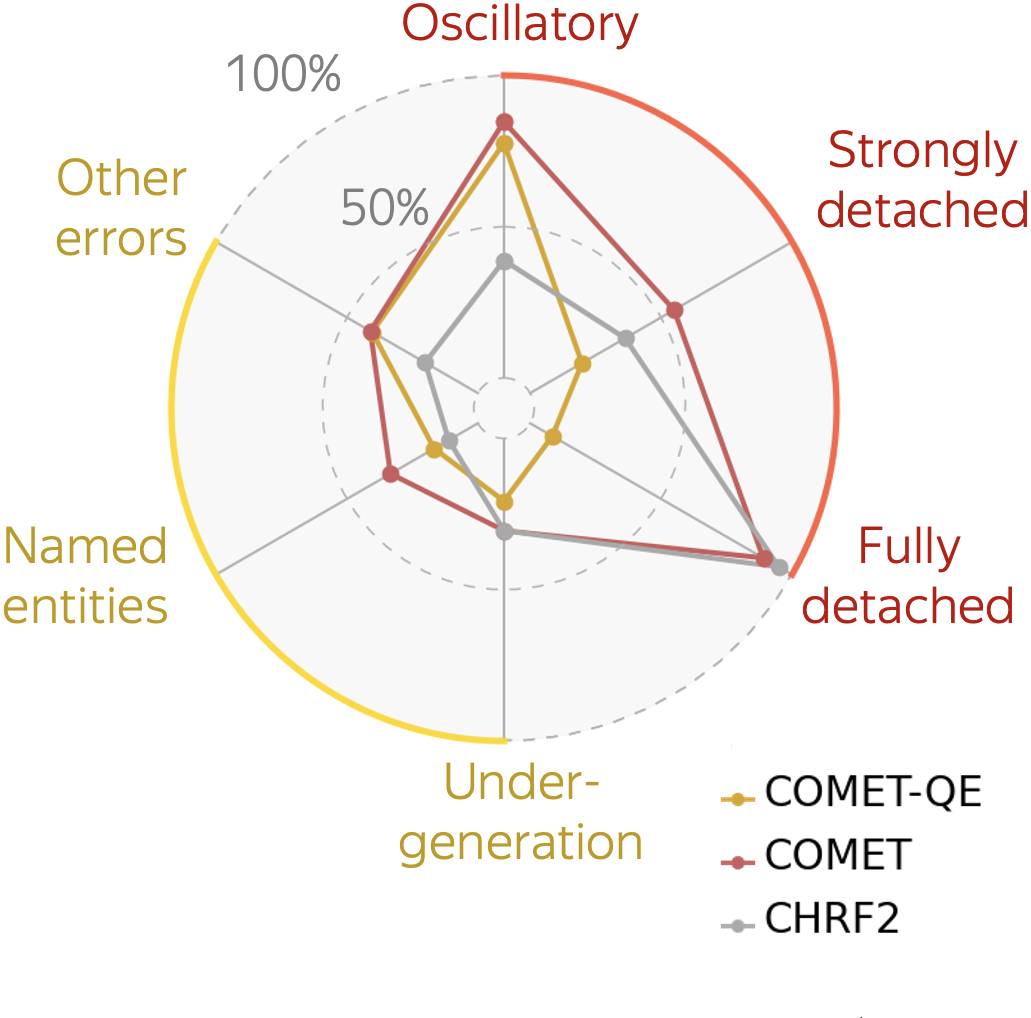}
        \caption{Quality filters}
        \label{fig:spiderC:filters}
    \end{subfigure}
    \caption{Distribution of the translations flagged by each method conditioned on each pathology.}
    \label{fig:spider_diagrams}
\end{figure*}

\subsection{Combining Heuristics and Quality Filters}
\label{subsec:lowqualfilter}

Intersecting a set of translations obtained via a heuristic with the bottom scored translations according to a quality filter was introduced in~\citet{raunak-etal-2021-curious}. The motivation for this was to avoid incorrectly flagging good translations as hallucinations (e.g., without this intersection, \textsf{RT} flags good-quality paraphrases). Intuitively, this idea is very reasonable: hallucinations are indeed incorrect translations. However, implementing this in practice requires a good quality estimation model, specifically for ranking poor translations. Unfortunately, we showed in  Section~\ref{subsec:qualityfilters} that for this purpose, even the state-of-the-art \textsf{COMET-QE} is largely inadequate. This means that using such quality estimates may lead to filtering out a lot of hallucinations, which is not desirable in preventive settings. Our results show that this is exactly the case: e.g., filtering with \textsf{COMET-QE} leads to losing nearly~$80\%$ of hallucinations detected by \textsf{Seq-Logprob}. All in all, such an intersection generally does more harm than good.

\section{Analysing Hallucination Pathologies}
\label{sec:mt_pathologies}

In this section, we look at hallucination pathologies
in isolation and show that the behavior of detection
methods varies depending on the type of pathology.
For example, for a given pathology, some methods
may be specialised, whereas others may fail. For a
similar analysis on other less severe translation errors,
refer to Appendix~\ref{sec_app:analysinglesssevere}.

\paragraph{Fully detached hallucinations.}
Figure~\ref{fig:spider_diagrams} shows that these hallucinations are easily detected by several methods, e.g. \textsf{Seq-Lobprob}, \textsf{Attn-ign-SRC}, \textsf{COMET}, \textsf{CHRF2}. This is not surprising: intuitively, the most severe pathology should be the easiest to detect. However, \textsf{COMET-QE} \textit{fails to identify almost all these hallucinations}. While \textsf{COMET}-based metrics are known to not penalise enough certain types of errors (e.g., discrepancies in numbers and named entities; see~\citet{chantal_2022,salted_raunak2022}), such poor performance for completely inadequate translations is highly unexpected. This calls for further research on the behavior of quality estimation models.

On a more general note, previous work suggested that fully detached hallucinations emerge as exact copies of references from training data~\cite{raunak-etal-2021-curious}. We validate this hypothesis and find the contrary: out of the 44 unique translations marked as fully detached from the source, only 4 are exact copies of references in the training data. Nevertheless, when looking at these sentences more closely, we see that they do contain large substrings that are seen frequently during training. Therefore, fully detached hallucinations are indeed likely to be traced back to the training data, but they \textit{emerge in non trivial ways and not necessarily as exact copies}. This can be seen as one more evidence that, when dealing with memorisation in language models, it is necessary to consider not just full copies in the training data but also near-duplicates~\cite{lee-etal-2022-deduplicating}.



\paragraph{Strongly detached hallucinations.}
As expected, this pathology is harder to detect than fully detached hallucinations~(Figure~\ref{fig:spider_diagrams}). However, the trends are largely similar: for example, \textsf{COMET-QE} fails again, and \textsf{Seq-Logprob} performs best and outperforms even reference-based \textsf{COMET}.

\paragraph{Oscillatory hallucinations.} The method specifically developed to detect this hallucination type~(Top $n$-gram count, \textsf{TNG}) performs worse than most of the other methods. Among the rest, \textsf{COMET} performs best. Interestingly, in contrast to previous observations, \textsf{COMET-QE} performs well, being on par with \textsf{COMET}.



\section{\textsc{DeHallucinator}: Overwriting Hallucinations at Test Time}
\label{sec:reversal}



In previous sections, we saw that hallucinations are more unstable than other translations: for them, generated MC-dropout hypotheses tend to vary greatly. This motivates us to look more closely into these hypotheses: are any of these translations \textit{not} hallucinations? Answering this question not only gives insight into the inner workings of hallucinating NMT models, but also leads to an interesting practical application~-- overwriting hallucinations at inference time. 
This is of utmost importance for production systems where hallucinations have a deeply compromising effect on user trust.

\begin{figure}[t]
\centering
{\includegraphics[width=\columnwidth]{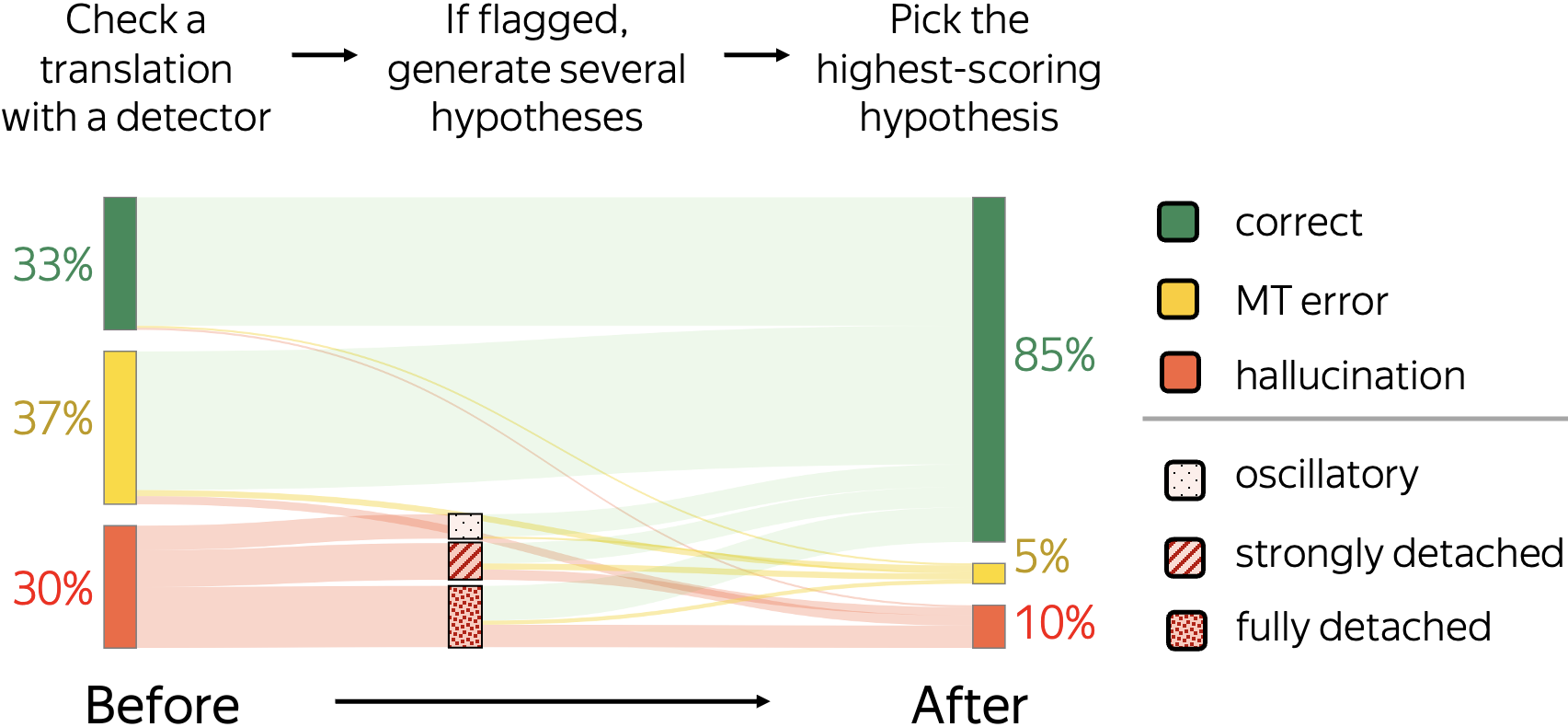}}
\caption{Our pipeline scheme along with results.}
\label{fig:reversal}
\end{figure}

\paragraph{Whenever flagged, overwrite with better.} Intuitively, our idea is similar to hybrid pipelines when a machine-generated translation is first passed to a quality estimation system and then, if needed, is corrected by human translators. In our case, we first apply a hallucination detector and then, if a translation is flagged, we try to overwrite it with a better translation~(Figure~\ref{fig:reversal}). For this, we generate several MC-dropout hypotheses, score them with some measure, and pick the highest-scoring translation as a final candidate (in the spirit of reranking approaches~\cite{shen-etal-2004-discriminative, lee-etal-2021-discriminative, fernandes-etal-2022-quality, freitag_decoding_2022}).

The general pipeline above relies on the choice of a hallucination detector and a scoring measure.
For the detector, we use the best of the analysed detectors, i.e. \textsf{Seq-Logprob}.\footnote{In this experiment, we take the translations from our dataset and consider the percentiles defined in Section~\ref{sec:halls_dataset}.} For the scoring measure, a natural choice would be a quality estimation system: by construction, these systems are designed to score translations according to quality. However, as we saw earlier, even the state-of-the-art \textsf{COMET-QE} may fail~(Section~\ref{subsec:qualityfilters}). Therefore, we compare two measures: \textsf{COMET-QE} and \textsf{Seq-Logprob}. 

In this experiment, we randomly choose 200~translations from our dataset flagged by \textsf{Seq-Logprob}. For each, we generate 10 hypotheses with MC-dropout. Then, for the overwritten translations we gather annotations according to our guidelines~(Section~\ref{sec:halls_dataset}). The results are summarized in Figure~\ref{fig:reversal}. Although we were concerned about \textsf{COMET-QE} because of its low performance when ranking poor translations, we find that for choosing the best hypothesis, it is indeed appropriate and performs better than \textsf{Seq-Logprob}. We thus show results with \textsf{COMET-QE} scores in the main text and with \textsf{Seq-Logprob} in Appendix~\ref{sec:overwriting_seqlogprob}.


\paragraph{Most hallucinations and errors become correct.} Figure~\ref{fig:reversal} shows that most hallucinations are overwritten with correct translations. This is surprising: in most cases, the model is not stuck in a hallucinatory mode and can generate good translations in a small vicinity of model parameters. In this sense, most hallucinations result from ``bad luck'' during generation and not profound model defect. Note that fully detached hallucinations are the hardest to improve. This makes sense as these are likely to be traced back to (near-)duplicates in the training data and, therefore, they do highlight model anomalies.

Note that in this pipeline, we overwrite not only hallucinations but also other errors and correct translations that were flagged by the detector. This means that our method needs to appropriately handle such translations. From Figure~\ref{fig:reversal} we see a pleasant side-effect: our approach overwrites most of the errors with correct translations. Just as importantly, almost all originally correct translations remain correct. Overall, the proportion of correct translations among the ones flagged by the detector increases from 33$\%$ to 85$\%$, and the hallucinatory rate decreases threefold. In Appendix~\ref{appendix:overwriting_examples}, we show several examples of overwritten hallucinations.

\section{Conclusions}

Dealing with hallucinations is difficult. First, we had to take a step back from previous work and refuse procedures that amplify the problem, as these hinder the behavior of models in their standard settings. After that, we notice that work on detection often relies on assumptions that remained unquestioned~(e.g., generic quality measures, targeted heuristics, anomalous attention being suitable detectors). Through extensive experiments, we establish order in detection methods. Surprisingly, despite introduction of several methods specifically targeted for hallucinations, what works best has always been at our disposal: standard sequence log-probability. This suggests that characteristics innate to a model can have a lot of value. In fact, such characteristics are the backbone of our \textsc{DeHallucinator}, a lightweight approach that significantly alleviates hallucinations at test time. This leaves space for future research on model uncertainty, hallucination prevention, understanding where hallucinations come from, among others. For this, we release our corpus with structured annotations along with the model and its training data. Altogether, this allows us to say that we provide solid ground for future study of hallucinations in~NMT.

\section*{Limitations}
We highlight three main limitations in our work. 
First, although the foundation of our proposed taxonomy for hallucinations rests on the idea of detachment from the source content, we do not evaluate it quantitatively. Indeed, we cannot point whether the model that generated the hallucinations was indeed detached from the source sentence when generating them. Nevertheless, we can guarantee that the hallucinations in our dataset are translations that are detached from the source content according to professional translators. We consider that the quantitative analysis of the detachment to be out of the scope of this paper. Still, it constitutes an interesting line for future research on understanding hallucinations in NMT that may be facilitated with the release of our code, model and annotated dataset.

Second, while this paper comprehensively studies the phenomena of hallucinations in NMT for a high-resource language pair, experiments in more language pairs (including low-resource languages) are necessary to assess the broad validity of our claims. To keep our setup familiar to researchers and practicioners, we opted for a familiar language pair for which data is widely available. Moreover our choice also facilitated the data collection process as there is a large supply of professional translators for this language pair. 

Lastly, instead of focusing on more recent NMT models that use large pretrained language models as their backbone, we focused on a Transformer base model. The reason for this choice is that we wanted to keep the setup simple, familiar, easy to reproduce, and computationally economical. Moreover, it was important for our work to have full control on the training and held-out data. Nevertheless, research on hallucinations on more recent and powerful NMT models is an exciting line of future work and we hope our work spurs that research.

\section*{Acknowledgments}
This work is partially supported by the European Research Council (ERC StG DeepSPIN 758969, by the FCT through contract UIDB/50008/2020, by EU’s Horizon Europe (UTTER, HORIZON-CL4-2021-HUMAN-01-13, contract 101070631), and by the P2020 programs MAIA and Unbabel4EU (LISBOA-01-0247-FEDER-045909 and LISBOA01-0247-FEDER-042671). Lena is supported by the Facebook PhD Fellowship.

\bibliography{custom, anthology}
\bibliographystyle{acl_natbib}

\newpage
\clearpage
\appendix

\section{Experimental setup}
\label{appendix:experimental_setup}
\paragraph{Data preprocessing.} We filter our data using language identification and simple length-heuristics described in \citet{tran-etal-2021-facebook}. We encode the data with byte-pair encoding~\citep{sennrich-etal-2016-neural} using the SentencePiece framework~\citep{kudo-richardson-2018-sentencepiece}. We set the vocabulary size to 32k and compute joint encodings and vocabulary. 

\paragraph{Model parameters.} We follow the setup of Transformer base model~\cite{transformer_vaswani}  (hidden size
of 512, feedforward size of 2048, 6 encoder and 6 decoder layers, 8 attention heads). The model has approximately 77M parameters.

\paragraph{Optimizer.} Similarly to~\cite{vashishthetal2019}, we train our model using the Adam optimizer with $\beta_1 = 0.9$ and $\beta_2 = 0.98$ and use an inverse square root learning rate scheduler with initial value $5 \times 10^{-4}$, and a linear warm-up in the first 4000 steps.

\paragraph{Training and Inference.} Models are trained for 250K updates with a batch size of about 32K tokens. We set dropout to 0.3. At inference time, we produce translations using beam search with a beam of 5. We validate our models during training using SacreBLEU \citep{post-2018-call}, and we choose the checkpoint based on best BLEU in validation. We provide BLEU\footnote{\textsf{BLEU+case.mixed+lang.XX+numrefs.1+\\smooth.exp+tok.13a+version.1.4.2}} and COMET baselines on WMT evaluation campaigns in Table~\ref{tab:appendix_baselines_wmt}. We train and performance inference on top of the Fairseq framework~\cite{ott-etal-2019-fairseq}.

\paragraph{COMET versions.} We use models available in the official repository\footnote{\textsf{https://github.com/Unbabel/COMET}}: \texttt{wmt20-comet-da} for COMET and \texttt{wmt20-comet-qe-da-v2} for COMET-QE.

\paragraph{\textbf{\textsf{TokHal-Model.}}} We follow the official implementation.\footnote{\textsf{https://github.com/violet-zct/fairseq-detect-hallucination}} For the synthetic data generation step, we used BART-large; and, for the token-level hallucination predictor, we used XLM-R~\cite{xlm_roberta_conneau}.

\paragraph{Computing Infrastructure.} All our experiments have been ran on a machine with 2 physical Intel(R) Xeon(R) Gold 6348 @ 2.60GHz CPUs (total of 112 threads), and 4 NVIDIA RTX A6000 GPUs. In particular, the NMT model described above was trained in less than 30 hours on a single GPU.



\section{Data Collection}
\label{appendix:data_collection}
We perform a rigorous and comprehensive manual annotation procedure with professional translators, in order to make sure that we are reliably analysing hallucinated translations.\footnote{Our translators were hired through Upwork and they were informed about the academic purposes of the data annotation process. All translators hired for this study reside in Europe.} First, the translators were asked to be familiar with our task by reading our provided annotation examples, along with detailed annotation instructions. Then they had to pass a test to show they can recognize different pathologies in different translations. We selected the two best translators to gather the annotations. After being hired, we ran three one-hour tutorial sessions to explain the task thoroughly and to clarify any possible questions. During the annotation process, we made sure to be promptly available to answer any question from the translators. We paid a fair wage  (25-30 USD per hour) --~well above both the US federal minimum and the average EU minimum wage~-- and inspected their work for quality. 

\paragraph{Guidelines.} We make available the full guidelines used by the translators along with all other resources in the project repository. In short, the annotators were presented with a source sentence and a model-generated hypothesis and asked to deem that translation as correct (\texttt{COR}) or incorrect. If incorrect, they were prompted to answer a series of yes/no questions, regarding the presence of specific hallucinatory pathologies: oscillations (\texttt{OSC}), strong detachment (\texttt{SD}) and full detachment (\texttt{FD}). We also asked annotators to flag critical errors such as named-entity mistranslations (\texttt{NE}) and under-generated translations (\texttt{UG}). 


\begin{table}[t]
\centering
\footnotesize
\renewcommand\arraystretch{1.15}
\resizebox{\columnwidth}{!}{
\begin{tabular}{l l l l}
\toprule
\textbf{Metric} & \textbf{WMT2014} & \textbf{WMT2017} & \textbf{WMT2018} \\\midrule
BLEU & 31.1 & 32.6 & 38.9 \\ 
COMET & 0.3178 & 0.3257 & 0.3340 \\ 
\bottomrule
\end{tabular}
}
\caption{Evaluation metrics for EN → DE for \textit{newstest} sets for the WMT 2014, 2017 and 2018 campaigns.}
\label{tab:appendix_baselines_wmt}
\end{table}

\begin{table}[t]
\centering
\footnotesize
\renewcommand\arraystretch{1.15}
\begin{tabular}{c c c c c c}
\toprule
\multicolumn{6}{c}{\textbf{Fleiss’s Kappa Scores}} \\
\texttt{COR} & \texttt{UG} & \texttt{NE} & \texttt{OSC} & \texttt{SD} & \texttt{FD} \\ \midrule
0.62 & 0.73 & 0.42 & 0.86 & 0.45 & 0.89\\ 
\bottomrule
\end{tabular}
\caption{Fleiss’s Kappa inter-annotator agreement scores (↑) for the different categories translators were prompted to identify.}
\label{tab:appendix_agreement}
\end{table}

\begin{figure*}[t]
\captionsetup[subfloat]{margin=10pt,format=hang,singlelinecheck=false,justification=centering}
    \centering
    \begin{subfigure}[b]{0.33\linewidth}        
        \centering
        \includegraphics[width=1\linewidth]{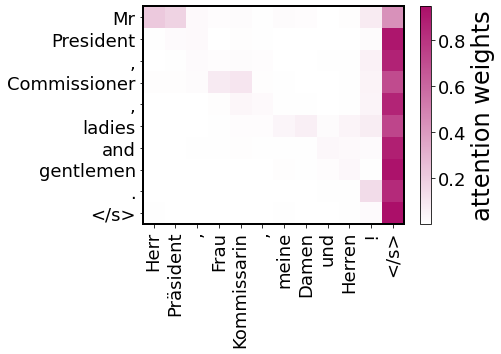}
        \caption{Correct translation,
        flagged by \textsf{Attn-to-EOS}}
        \label{fig:attn_correct}
    \end{subfigure}
    \ \ 
    \begin{subfigure}[b]{0.28\linewidth}        
        \centering
        \includegraphics[width=1\linewidth]{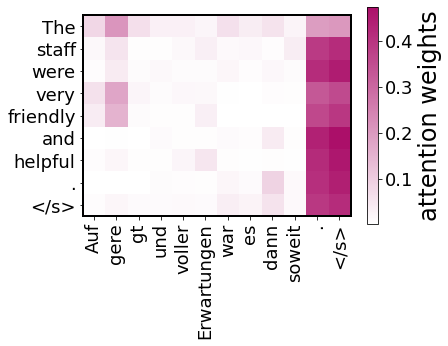}
        \caption{Hallucination,
        flagged by \textsf{Attn-ign-SRC}}
        \label{fig:attn_strange}
    \end{subfigure}
    \ \ 
    \begin{subfigure}[b]{0.35\linewidth}        
        \centering
        \includegraphics[width=1\linewidth]{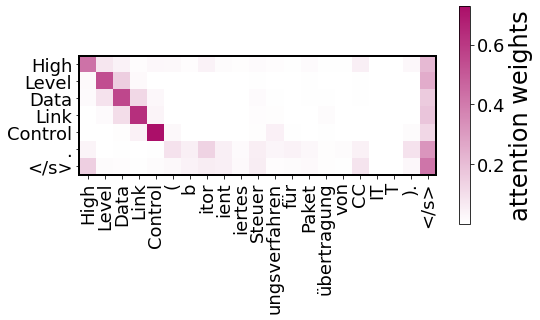}
        \caption{Undergeneration,
        flagged by \textsf{Attn-ign-SRC}}
        \label{fig:attn_omission}
    \end{subfigure}
    \caption{Examples of attention maps flagged by attention-based heuristics.}
    \label{fig:attn_patterns}
\end{figure*}


\paragraph{Inter-annotator agreement.} To determine the reliability of our annotations, we asked both our translators to annotate a set of 400 randomly sampled translations. For all hallucinatory categories but \texttt{SD}, the annotators achieved -- according to Cohen's kappa coefficient \citep{kappa_cohen} --  almost perfect agreement. For all other categories, moderate to substantial agreement was obtained. This confirms that our data conforms very well to our instructions. The agreement scores for each category are displayed in Table~\ref{tab:appendix_agreement}.

\section{Patterns of attention maps for translations flagged with attention-based heuristics}
\label{app:attn_patterns}
The attention maps are shown in Figure \ref{fig:attn_patterns}. While both \textsf{Attn-to-EOS} and \textsf{Attn-ign-SRC} were designed to identify translations for which almost all the attention mass is concentrated on the \texttt{EOS} token, the patterns identified with \textsf{Attn-ign-SRC} are more diverse. They span from attention mass coming to various uninformative tokens (e.g., punctuation and other tokens as in Figure~\ref{fig:attn_strange}) to examples shown in Figure~\ref{fig:attn_omission} where attention is
mostly diagonal~(typically, these correspond to undergenerations; we follow this discussion in the next section).

\section{Analysing Less Severe Translation Errors}
\label{sec_app:analysinglesssevere}
We notice that some of the detection methods are specialised on specific pathologies. For example, \textsf{Attn-ign-SRC} is by far the best for detecting undegenerations. This is expected: an undergeneration does not cover part of the source sentence, thus a significant proportion of source tokens receives little attention mass~(see example in Figure~\ref{fig:attn_omission}). For named entity errors, the best heuristic is~$\textsf{TokHal-Model}$. This mirrors the discussion in Section~\ref{subsec:detection_heuristics}: 
while severe errors (i.e., hallucinations) fall out of distribution for this model, mistranslations of short phrases are more in line with the model's training.

On a different note, $\textsf{MC-DSim}$ performs much better for hallucinations than for less severe pathologies~(Figure~\ref{fig:spiderB:heuristics_continuous}). This again points to hallucinations being more unstable than other errors.


\section{Overwriting Hallucinations with \textsf{Seq-Logprob} as the scoring measure}
\label{sec:overwriting_seqlogprob}
\begin{figure}[t]
\centering
{\includegraphics[width=\columnwidth]{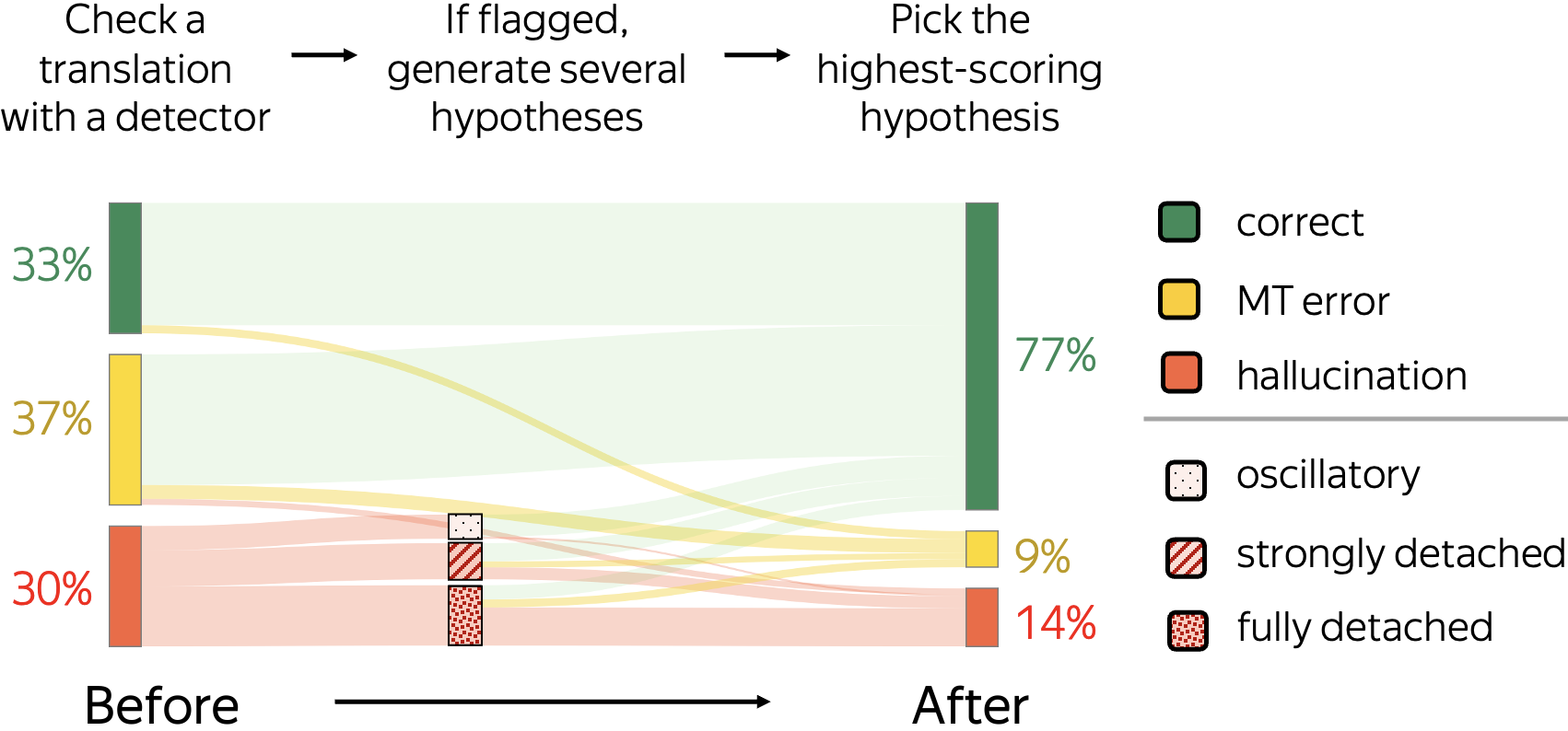}}
\caption{Our pipeline scheme along with results when we use \textsf{Seq-Logprob} as both the detector and the scoring measure.}
\label{fig:reversal_seqlogprob}
\end{figure}

The results are shown in Figure~\ref{fig:reversal_seqlogprob}. In order to use \textsf{Seq-Logprob} as the scoring measure, we score all generated hypothesis with the original model. Overall, although the results follow the same trend as those using \textsf{COMET-QE} as the scoring measure, they are slightly worse: the hallucinatory rate is higher and the percentage of correct translations is smaller.

\section{Examples of overwritten translations}
\label{appendix:overwriting_examples}
Table~\ref{tab:overwriting_hallucinations} shows examples of each hallucination type that have been overwritten with correct translations with the approach described in Section~\ref{sec:reversal}.

\section{Practical Recommendations for Detection}
\label{sec:prac_recommendations}
As we have mentioned in Section~\ref{subsec:binary_cont_scores}, Binary-label heuristics output a value in $\{0; 1\}$, and continuous-score heuristics output a value $s$ in $\mathbb{R}$. These values $s$ can be used to build a binary decision rule: for a given source $x$, translation $\hat{y}$ and reference $y$, a translation is flagged by the detector if and only if $s(x, \hat{y}, y) \leq \gamma$. Naturally, $s$ may not need be a function of $x$, $\hat{y}$ and $y$ (e.g. \textsf{COMET-QE} is only a function of $x$ and $\hat{y}$).
\paragraph{Choosing the thresholds.} In our work, we chose the thresholds $\gamma$ for each detector by assessing the value $s$ correspondent to approximately the $0.4$-th percentile to be consistent with the data selection process (see Section~\ref{sec:halls_dataset}). We computed these thresholds using the entire filtered held-out data. However, in practice, we obtained very similar results when we obtained these thresholds using a collection of only 10\,000 examples from that dataset.

We recommend computing these thresholds on in-domain clean data. This will guarantee that the cut-off values were obtained in a scenario where the model performs best. Finally, the definition of the $k$-th percentile is expected to have an impact on the precision-recall trade-off. Thus, for preventive settings, we recommend sticking to more conservative values of $k$. 


\paragraph{Choosing the detectors.} The choice of detector rests upon the application for which it is intended. For high-precision settings, binary-label heuristics such as $\textsf{TNG}$ and variants thereof~\cite{salted_raunak2022} may be more recommended. For preventive settings, we suggest using \textsf{Seq-Logprob} as the backbone of the hallucination detector. Naturally, we generally obtain higher recall by joining the sets of translations flagged with multiple methods. For example, Figure~\ref{fig:spider_diagrams} reveals that joining the set of translations flagged with \textsf{Seq-Logprob} with the set of translations flagged with \textsf{COMET-QE} is very reasonable: \textsf{COMET-QE} performs better for oscillatory hallucinations, while \textsf{Seq-Logprob} is better for the other hallucination types.

\paragraph{Be careful when relying on references.} Using reference information is helpful for detecting hallucinations~(see Section~\ref{subsec:qualityfilters}), and while it may not be used to detect hallucinatins on-the-fly, it may still prove useful for analysis works. We have found that high-quality parallel data is critical for adequate application of these methods: very low scores might not only be attributed to poor translations, but also to reference mismatches. Indeed, preliminary experiments highlighted this worrying trend, which motivated us to clean the held-out set (Section~\ref{sec:experimental_setting}). Thus, if using reference information to detect hallucinations, make sure to thoroughly clean your parallel data.

\definecolor{GoodGreenText}{HTML}{397e64}
\definecolor{GoodGreenShade}{HTML}{e9f9f5}
\DeclareRobustCommand{\hlgreen}[1]{{\sethlcolor{GoodGreenShade}\hl{#1}}}

\begin{table*}
    \footnotesize
    \centering
    \renewcommand\arraystretch{1.15}
    \begin{tabular}{>{\arraybackslash}m{0.15\textwidth} >{\arraybackslash}m{0.8\textwidth}}
    \toprule
        \multicolumn{2}{c}{\textsc{\textbf{Overwriting fully detached hallucinations}}} \\ \midrule
        \textsc{Source} &  Handys, die bis auf Wäschewaschen und Staubsaugen scheinbar alles können.\\
        \textsc{Reference} & Mobile phones that can practically do everything except clean the laundry and vacuum clean.\\ \arrayrulecolor{gray!30}\midrule
        \textsc{Original \,\,\, \,\,\, \,\,\, Hypothesis} & \textcolor{HallRedText}{\hl{The staff were very friendly and helpful.}}\\ 
        \textsc{Overwritten \,\,\, Hypothesis} & \textcolor{GoodGreenText}{\hlgreen{Mobile phones that seem to be able to do everything except on laundry and dustproofing.}}\\ \arrayrulecolor{gray!80}\bottomrule
    \end{tabular}\vspace{5pt}
    
    \begin{tabular}{>{\arraybackslash}m{0.15\textwidth} >{\arraybackslash}m{0.8\textwidth}}
        \textsc{Source} &  In unserem 2 Personen Van mit Dusche/WC war ausreichend Platz für uns beide.\\
        \textsc{Reference} & The space in our 2 person van with shower/toilet was enough for 2 people.\\ \arrayrulecolor{gray!30}\midrule
        \textsc{Original \,\,\, \,\,\, \,\,\, Hypothesis} & \textcolor{HallRedText}{\hl{The staff were very friendly and helpful. The room was clean and comfortable.}}\\ 
        \textsc{Overwritten \,\,\, Hypothesis} & \textcolor{GoodGreenText}{\hlgreen{In our 2 person van with shower/WC there was enough room for us both.}}\\ \arrayrulecolor{black}\bottomrule
    \end{tabular}\vspace{15pt}

    \renewcommand\arraystretch{1.15}
    \begin{tabular}{>{\arraybackslash}m{0.15\textwidth} >{\arraybackslash}m{0.8\textwidth}}
    \toprule
        \multicolumn{2}{c}{\textsc{\textbf{Overwriting strongly detached hallucinations}}} \\ \midrule
        \textsc{Source} &  Tickets für Busse und die U-Bahn ist zu teuer, vor allem in Stockholm.\\
        \textsc{Reference} & Tickets for buses and the subway is too expensive, especially in Stockholm.\\ \arrayrulecolor{gray!30}\midrule
        \textsc{Original \,\,\, \,\,\, \,\,\, Hypothesis} & \textcolor{HallRedText}{\hl{The hotel is located in the centre of}} Stockholm, \textcolor{HallRedText}{\hl{close to the}} train \textcolor{HallRedText}{\hl{station}}.\\ 
        \textsc{Overwritten \,\,\, Hypothesis} & \textcolor{GoodGreenText}{\hlgreen{Buses and metro tickets are too expensive, especially in Stockholm.}}\\ \arrayrulecolor{gray!80}\bottomrule
    \end{tabular}\vspace{5pt}
    
    \begin{tabular}{>{\arraybackslash}m{0.15\textwidth} >{\arraybackslash}m{0.8\textwidth}}
        \textsc{Source} &  Ich freue mich über jeden Tag, wo es mir so gut geht!\\
        \textsc{Reference} & I am pleased about each day, where I am so well!\\ \arrayrulecolor{gray!30}\midrule
        \textsc{Original \,\,\, \,\,\, \,\,\, Hypothesis} & I \textcolor{HallRedText}{\hl{look forward to seeing you}} every day!\\ 
        \textsc{Overwritten \,\,\, Hypothesis} & \textcolor{GoodGreenText}{\hlgreen{I'm very happy with every day I'm doing so well!}}\\ \arrayrulecolor{black}\bottomrule
    \end{tabular}\vspace{15pt}

    \begin{tabular}{>{\arraybackslash}m{0.15\textwidth} >{\arraybackslash}m{0.8\textwidth}}
    \toprule
        \multicolumn{2}{c}{\textsc{\textbf{Overwriting oscillatory hallucinations}}} \\ \midrule
        \textsc{Source} &  In dieser Zeit stürzt sich Murnau bereits wieder in Theaterproben, allerdings widmet er sich nicht mehr der Schauspielerei, sondern der Regie.\\
        \textsc{Reference} & During this period, Murnau once again dedicates his time to theatre rehearsals; however, this time not as an actor, but as a director.\\ \arrayrulecolor{gray!30}\midrule
        \textsc{Original \,\,\, \,\,\, \,\,\, Hypothesis} & Murnau \textcolor{HallRedText}{\hl{was born in Murnau, Germany}}. He \textcolor{HallRedText}{\hl{was born in Murnau, Germany.}}\\ 
        \textsc{Overwritten \,\,\, Hypothesis} & \textcolor{GoodGreenText}{\hlgreen{During this time, Murnau began to appear in theater tests again, but he was no longer concerned with acting, but with directing.}}\\ \arrayrulecolor{gray!80}\bottomrule
    \end{tabular}\vspace{5pt}
    
    \begin{tabular}{>{\arraybackslash}m{0.15\textwidth} >{\arraybackslash}m{0.8\textwidth}}
        \textsc{Source} &  Die Staaten, deren Fangflotten im Nordwestatlantik Hochseefischerei betreiben, bemühen sich im Rahmen der NAFO (Organisation für die Fischerei im Nordwestatlantik) um eine gemeinsame Bestandserhaltung und - bewirtschaftung.\\
        \textsc{Reference} & States which fish in the high seas in the North West Atlantic co-operate in NAFO (North-west Atlantic Fisheries Organisation) in order to ensure conservation and management of stocks.\\ \arrayrulecolor{gray!30}\midrule
        \textsc{Original \,\,\, \,\,\, \,\,\, Hypothesis} & The \textcolor{HallRedText}{\hl{North-West Atlantic Fisheries Organisation (NAFO)}} is a member of the \textcolor{HallRedText}{\hl{North-West Atlantic Fisheries Organisation (NAFO)}}.\\ 
        \textsc{Overwritten \,\,\, Hypothesis} & \textcolor{GoodGreenText}{\hlgreen{The states whose fishing fleets in the North-West Atlantic are engaged in deep-sea fishing are seeking joint conservation and management within the framework of the North-West Atlantic Fisheries Organisation (NAFO).}}\\ \arrayrulecolor{black}\bottomrule
    \end{tabular}
    \caption{Examples of hallucinations of each type that have been overwritten with correct translations.}
    \label{tab:overwriting_hallucinations}
\end{table*}





\end{document}